\documentclass[sigconf]{acmart}

\def\BibTeX{{\rm B\kern-.05em{\sc i\kern-.025em b}\kern-.08emT\kern-.1667em\lower.7ex\hbox{E}\kern-.125emX}}

\setcopyright{rightsretained} 
\copyrightyear{2019} 
\acmYear{2019} 
\acmConference[AIES '19]{AAAI/ACM Conference on AI, Ethics, and Society}{January 27--28, 2019}{Honolulu, HI, USA}
\acmBooktitle{AAAI/ACM Conference on AI, Ethics, and Society (AIES '19), January 27--28, 2019, Honolulu, HI, USA}
\acmDOI{10.1145/3306618.3314266}
\acmISBN{978-1-4503-6324-2/19/01}
\acmPrice{}

\usepackage{Definitions}
\usepackage[utf8]{inputenc}

\usepackage{epstopdf}
\usepackage{subfig} 
\usepackage[export]{adjustbox}
\usepackage{booktabs}       
\begin{document}

\title{Loss-Aversively Fair Classification}

\author{Junaid Ali, Muhammad Bilal Zafar, Adish Singla, Krishna P. Gummadi }

\affiliation{
  \institution{Max Planck Institute for Software Systems (MPI-SWS)}
          \city{\{junaid, mzafar, adishs, gummadi\}@mpi-sws.org}
}
\renewcommand{\shortauthors}{Ali, et al.}



\newcommand{\todo}[1]{\textcolor{red}{Page limit: #1}}
\newcommand{\junaid}[1]{\textcolor{pink}{Junaid: #1}}
\newcommand{\bilal}[1]{\textcolor{blue}{Bilal: #1}}
\newcommand{\krishna}[1]{\textcolor{olive}{Krishna: #1}}
\newcommand{\xhdr}[1]{\vspace{0.2mm}\noindent{{\bf #1.}}}
\newcommand{\xytup}{\xb,y}
\newtheorem{prop}{Proposition}
\newtheorem{claim}{Claim}
\newcommand{\dt}{disparate treatment}
\newcommand{\di}{disparate impact}
\newcommand{\dm}{disparate mistreatment}
\newcommand{\stp}{statistical parity}
\newcommand{\eop}{equality of opportunity}

\begin{abstract} 

The use of algorithmic (learning-based) decision making in scenarios
that affect human lives has motivated a number of recent studies to
investigate such decision making systems for potential unfairness,
such as discrimination against subjects based on their sensitive
features like gender or race.
However, when judging the fairness of a newly designed decision making
system, these studies have overlooked an important influence on
people's perceptions of fairness, which is how the new algorithm
changes the status quo, i.e., decisions of the existing decision
making system.
Motivated by extensive literature in behavioral economics and
behavioral psychology (prospect theory),
we propose a notion of fair updates that we refer to as {\it
loss-averse updates}. Loss-averse updates constrain the updates to yield improved (more beneficial) outcomes to subjects compared to the status quo.
 We propose tractable proxy measures that would
allow this notion to be incorporated in the training of a variety of
linear and non-linear classifiers.  We show how our proxy measures can
be combined with existing measures for training nondiscriminatory
classifiers.
Our evaluation using synthetic and real-world datasets demonstrates that the proposed proxy measures are effective for their desired tasks.

\end{abstract}

\maketitle

\section {Introduction}\label{sec:intro}

The use of algorithmic (data-driven and learning-based) decision
making systems in domains ranging from judiciary (recidivism risk
estimation) and banking (credit ratings and loan approval risk) to
welfare (benefits eligibility) and insurance (accident risks) has
raised numerous concerns about their fairness.
Consequently, in recent years, a number of notions of algorithmic
(un)fairness have been
proposed~\cite{pedreschi_discrimination,zafar_dmt,hardt_nips16} and
numerous learning mechanisms have been devised to train algorithmic
decision making systems that satisfy these
notions~\cite{Dwork2012,zafar_fairness,zafar_dmt,hardt_nips16,feldman_kdd15,icml2013_zemel13,pedreschi_discrimination,bechavod_fatml}. These
fairness notions have focussed on both the {\it decision outcomes} as
well as the {\it decision making process}, i.e., the inputs used to
make the decisions and the objectives of the learning algorithms.

In this paper, we focus on a crucial aspect of algorithmic decision
making systems ignored by existing studies on fair learning namely,
fairness of {\it updates to decision making systems}. 
In many decision making scenarios such as banking or judiciary or
insurance, a newly deployed system replaces an already existing
decision making system, be it run by a human decision maker or an
older learning model (e.g., learning models without
discrimination-awareness) or a learning model trained over outdated
training data (e.g., when features of users in a society evolve).
Existing literature in behavioral economics and psychology shows that
peoples' perceptions of fairness of the new decision making system are
influenced by {\it how the decision outcomes change from the status
  quo} i.e., how the new outcomes differ from the old
outcomes~\cite{kahneman1986fairness,Kahneman79prospecttheory,urbany1989all,bazerman1995perceptions}.
However, current works on fair learning do not account for the status
quo when reasoning about fairness of a decision making system.

In this work, inspired by existing literature in behavioral economics,
we formally define a notion of update fairness namely, {\bf
  loss-aversively fair updates}. Intuitively, our notion of loss-averse updates accounts
for the ``endowment effect'' in human behavior~\cite{kahneman1986fairness,Kahneman79prospecttheory}, where an
individual or a group of users perceives the fairness of the new
system based on whether their new outcomes were more or less
beneficial than their status quo outcomes from the existing
system. 

We design intuitive measures for this notion that can be
incorporated into a variety of linear and non-linear classifiers as
convex constraints and be efficiently learned. A classifier trained
using our constraints would account for the existing outcomes from the
status quo classifier.

We also show that our new notion of fair update can be easily
integrated with existing mechanisms for training non-discriminatory
classifiers. For instance, when attempting to equalize rates of
beneficial outcomes such as positive class acceptance rate or true
positive rate across different groups, adding our loss-averse update
constraint ensures that ``no group of users is worse-off'' than
before. Such a constraint may be necessary in practice when training
non-discriminatory classifiers as Bazerman et
al.~\cite{bazerman1995perceptions} point out that same ``don't make
anyone worse off' principle likely underlines Supreme Courts
decision~\cite{martin_scotus} that firing personnel from historically
advantaged groups to achieve parity (in order to overcome past
discrimination) is prohibited.

In the rest of the paper, we first formally define our notion of fair
update in the context of training classifiers.  We also propose
tractable and efficient mechanisms to train fair classifiers while
satisfying this practical consideration.
Experiments with synthetic and real-world datasets show the effectiveness of our mechanism in enforcing this consideration.

\section {Related work}\label{sec:related}
\xhdr{Fairness in ML}
A plethora of recent studies have focused on proposing
notions~\cite{zafar_dmt,hardt_nips16,pedreschi_discrimination,zafar_preferred}
and mechanisms for fairness-aware
classification~\cite{Dwork2012,zafar_fairness,zafar_dmt,hardt_nips16,feldman_kdd15,icml2013_zemel13,pedreschi_discrimination,bechavod_fatml,goel_cost_fairness,zafar_preferred}. For
more discussion into these notions, we point the interested readers
to~\cite{barocas_2016,berk2017fairness,zafar_fairness,salvatore_survey}. While
classification has received most attention in the area of
fairness-aware machine learning, some recent work has also focused on
prediction tasks beyond classification, such as
regression~\cite{berk2017convex},
ranking~\cite{singh2018fairness,beiga_ranking} and
clustering~\cite{fairlets_clustering}.
In this paper, we primarily focus on updates to
classification tasks, leaving fairness of updates to regression,
ranking, and clustering tasks to future studies.

\xhdr{Indvidual-level vs. Group-level Fairness Notions}
Fairness in classification has been divided into two broad areas:
individual- and group-level fairness~\cite{Dwork2012}. Loss-averse updates can be applied at both individual and group-levels. However, in this work, we only show results at the group-level. 

\xhdr{Normative vs. Descriptive Notions of Fairness} Our
fairness consideration for updating decision making systems has
roots in normative vs. descriptive approaches in behavioral
economics~\cite{Kahneman79prospecttheory,kahneman1986fairness}.  For
example, Kahneman et al.~\cite{kahneman1986fairness} show how certain
changes to an economic model that are accepted on the normative
standards might be deemed unacceptable on the descriptive standards.
Our work here is motivated by such observations: while
anti-discrimination laws (normatively) prescribe how
nondiscriminatory decisions ought to be done, if people
(descriptively) preceived the changes in outcomes with the new
nondiscriminatory decision system to be too disruptively
disadvantageous to them, they would resist adopting the new
system. Our notions of update fairness can be thought of as addressing
such practical considerations.

\section {Formalizing Notion of Loss-Averse Updates}\label{sec:setup}
\noindent In this section, we formally define a notion of fairness
that can be useful when updating algorithmic decision making systems.
Specifically, we focus on decision making tasks centered around binary
classification.

\xhdr{Preliminaries} In a binary classification task, given a training
dataset $\mathcal{D} = \{ (\xb_i, y_i) \}_{i=1}^{N}$, the goal is to
learn a function $\thetab: \RR^d \to \{-1,1\}$ between the feature
vectors $\xb \in \RR^d$ and class labels $y \in \{-1, 1\}$. For
convex decision boundary-based classifiers like logistic regression
and (non)linear SVM, this task boils down to finding a decision
boundary $\thetab^*$ in the feature space that minimizes a given loss
$L(\thetab)$ over $\Dcal$, \ie, $\thetab^{*} = \argmin_{\thetab}
L(\thetab)$.
The convexity of the loss function ensures that the optimal decision boundary parameters can be found in an efficient manner.
Then, for a given (potentially unseen) feature vector $\xb$, 
one predicts the class label 
$\hat{y} = 1$ if $d_{\thetab^{*}}(\xb) \geq 0$ and 
$\hat{y} = -1$ 
otherwise, where $d_{\thetab^{*}}(\xb)$ denotes the signed distance from 
$\xb$ to the decision boundary. 
Without loss of generality, we consider $\hat{y}=1$
to be the beneficial (desired) label, \eg, being granted the loan or being released on bail.

\xhdr{Setup} 
We consider scenarios where we need to update an
existing, {\it status quo}, binary classifier, whose decision boundary
is denoted by $\thetab_{sqo}$. We assume that the boundary of the new
classifier, $\thetab_{new}$ is learnt from the training dataset
$\mathcal{D}$. The outcomes of the updated (new) classifier may differ
from the status quo for many reasons such as the status quo classifier
being a human or an older (simpler) learning model, or the status quo
classifier being trained on out-dated training data, or the status quo
classifier being trained using models without awareness of potential
for discrimination. Our notion of fair update defines the conditions
in which the {\it changes in decision outcomes caused by an update}
would be deemed as fair.

\xhdr{Existing Notions: Discrimination in Classification}\\
Anti-discrimination laws require classification outcomes are also
required to be nondiscriminatory with respect to a sensitive feature
$z \in \{0,1\}$, \eg, gender, race.  Most of the existing studies
differentiate between the following two notions of discrimination:
\emph{statistical parity}~\cite{feldman_kdd15,Dwork2012}---also
referred to as \di, and \emph{equality of
  opportunity}~\cite{zafar_dmt,hardt_nips16}---also referred to as
\dm. Both notions require that certain group-conditional beneficial
outcome rates be the same for each group,
\ie,:
\begin{equation}\label{eq:nondiscrimination_cond}
\Bcal_{z=0}(\thetab)=\Bcal_{z=1}(\thetab),
\end{equation}
where the definition of the benefit function $\Bcal_z$ depends on the  notion of discrimination under consideration.

Under the notion of \emph{statistical
parity (SP)}~\cite{Dwork2012,feldman_kdd15}, the benefits function is defined as the positive class acceptance rate (AR), \ie,  the positive class acceptance rate should be the same for both the groups. More formally,
\begin{align}\label{eq:sp_nonconvex}
\text{---} \textit{\textbf{ SP:}} \qquad \qquad
P(\hat{y} = 1 | z = 0 ) = P(\hat{y} = 1 | z = 1 ), 
\end{align}

Under \emph{equality of opportunity (EOP)}
notion~\cite{hardt_nips16,zafar_dmt}, the benefit function is defined as the true positive rate, i.e., the true positive rate (TPR) should be
the same for both the groups. More formally, 
\begin{align}\label{eq:eop_nonconvex}
\text{---} \textit{\textbf{ EOP:}} ~ 
P(\hat{y} = 1 | y = 1, z = 0 ) = P(\hat{y} = 1 | y = 1, z = 1 ), 
\end{align}
Note that, current notions of nondiscrimination do not take into account status quo classifier. In the following section we introduce a notion of updating status quo classifier.

\xhdr{New Notion: Loss-Averse Updates}
We now formally describe a new consideration of fair updates, introduced in Section~\ref{sec:intro}. We draw inspiration from human behavior and behavioral economics and we consider how people might perceive fairness of an updated classifier in comparison to status quo. Specifically, any disadvantageous effect of an updated classifier would be considered unfair. Prospect theory, proposed by \citet{Kahneman79prospecttheory},  states that equal amounts of loses result in a bigger loss in utility than the increase in utility by the same amount of gains. In other words people percieve losses much worse than gains, \ie, they are loss-averse. Given the {\it status quo} classifier $\thetab_{sqo}$, a new classifier $\thetab_{new}$ constitutes a {\it loss-averse} update only when the new classifier increases the beneficial outcome rates for all groups. More formally,
\begin{align}
\Bcal_{z = k }(\thetab_{new}) \ \ \geq\ \  \Bcal_{z = k}(\thetab_{sqo}), \quad \text{for all } k \in \{0,1\} \label{eq:loss_averse_cons}
\end{align}
where $\Bcal_z$ can be any one of the benefit functions proposed in
the existing literature on nondiscriminatory classification.

\section {Updating Classifiers Loss-Aversively} \label{sec:methodology}
In this section, we devise mechanisms to update status quo classifier, $\thetab_{sqo}$ to $\thetab_{new}$ that follow the practical considerations of ``loss-averse updates''. We specifically focus on training convex decision boundary based classifiers (\eg, logistic regression, linear and non-linear SVMs), \ie, the classifiers that learn the decision boundary parameters by optimizing a convex loss function $L(\thetab)$.

\xhdr{Existing Mechanisms: Nondiscriminatory Classification}
Existing mechanisms to train nondiscriminatory classifiers involve  solving an optimization problem maximizing accuracy while equalizing benefits, \ie, enforcing Eq.~\eqref{eq:nondiscrimination_cond}, for different sensitive feature groups. More formally, 
\begin{align}\label{eq:nondiscriminatory_clf}
\mbox{minimize}\quad &   L(\thetab) \tag{P1}   \\ \nonumber
\mbox{subject to}\quad & \Bcal_{z=0}(\thetab)=\Bcal_{z=1}(\thetab) \nonumber,
\end{align}
Constraints in Problem~\eqref{eq:nondiscriminatory_clf}, as operationalized in Eqs.~\eqref{eq:sp_nonconvex} and \eqref{eq:eop_nonconvex} are non-convex. However, prior studies~\cite{zafar_fairness,zafar_dmt,bechavod_fatml} propose tractable convex or convex-concave proxies for enforcing the equality of benefits constraint in Eqs.~\eqref{eq:sp_nonconvex} and ~\eqref{eq:eop_nonconvex}. Borrowing these proxies from~\cite{zafar_fairness,zafar_dmt,bechavod_fatml}, one can replace the equal benefits condition with proxies as follows:
\begin{align}\label{eq:sp_cvx}
\text{---} \textit{\textbf{ SP:}}
\qquad \qquad 
\frac{1}{|\Dcal|}
\bigg | 
\sum_{(\xb,z) \in \Dcal}  (z - \bar{z}) d_{\thetab}(\xb_i) 
\bigg| 
\leq c,
\end{align}
\begin{align}\label{eq:eop_cvx}
\text{---} \textit{\textbf{ EOP:}}
\qquad 
\frac{1}{|\Dcal_+|} 
\bigg | 
\sum_{(\xb,z) \in \Dcal_+}  (z - \bar{z}) d_{\thetab}(\xb_i) 
\bigg| 
\leq c,
\end{align}
where $\Dcal_+$ are data points with  $y = 1$. Here \eop{}  limits discrimination in true positive rates of different groups. The covariance threshold $c\in\RR^+$ determines the level of discrimination, with $c=0$ aiming for a perfectly fair classifier.

\xhdr{New Mechanism: Loss-Averse Updates}
For updating the status quo classifier, $\thetab_{sqo}$, in a nondiscriminatory and loss-aversive manner, one can add the respective conditions to the classifier formulation as a constraint, \ie,
\begin{align}\label{eq:better_off_non_cvx}
\mbox{minimize}\quad &   L(\thetab) \tag{P2} \\ \nonumber
\mbox{subject to}\quad & \Bcal_{z=0}(\thetab)=\Bcal_{z=1}(\thetab) \\ \nonumber 
\quad & \bm{\Bcal_{z = k}(\thetab) \geq \Bcal_{z = k}(\thetab_{sqo})} , \quad \text{for all } k \in \{0,1\}. \nonumber
\end{align}
The constraints in the above problem are nonconvex functions of the classifier parameters $\thetab$, if $\Bcal$ is defined in terms of probabilities as given in Eqs.~\eqref{eq:sp_nonconvex} and \eqref{eq:eop_nonconvex}, for example, this would make it very challenging to solve the resulting problem in an efficient manner.

We used the convex proxies from prior studies ~\cite{zafar_fairness,zafar_dmt,bechavod_fatml} for the first constraint as given by Eqs.~\eqref{eq:sp_cvx} and ~\eqref{eq:eop_cvx}. We propose the following convex proxies to approximate the new loss-averse constraints in Problem~\eqref{eq:better_off_non_cvx}:

\noindent
Under SP, when the benefit function is AR we suggest:
\begin{align}\label{eq:better_off_more_ben_cvx_ar}
&\frac{1}{|\Dcal_{z=k}|} \sum_{\xb \in \Dcal_{z=k}} d_{\thetab}(\xb) \geq \frac{1}{|\Dcal_{z=k}|} \sum_{\xb \in \Dcal_{z=k}} d_{\thetab_{sqo}}(\xb) + \gamma, \qquad\\ \nonumber
&\quad \text{for all } k \in \{0,1\}, \gamma\in\RR^+.
\end{align}
Under EOP, when the benefit function is TPR we suggest:
\begin{align}\label{eq:better_off_more_ben_cvx_tpr}
&\frac{1}{|\Dcal_{z=k}^+|} \sum_{\xb \in \Dcal_{z=k}^+} d_{\thetab}(\xb) \geq \frac{1}{|\Dcal_{z=k}^+|} \sum_{\xb \in \Dcal_{z=k}^+} d_{\thetab_{sqo}}(\xb) + \gamma, \qquad\\ \nonumber
&\quad \text{for all } k \in \{0,1\}, \gamma\in\RR^+,
\end{align}
where $\Dcal_{z=k}$ are the data points whose sensitive attribute value $z = k$, and $\Dcal_{z=k}^+$ are data points in the dataset with label $y = 1$ and sensitive attribute value $z = k$. Here, $\gamma$ controls the strength of the constraint. We pick an appropriate $\gamma$ using a validation set. Note that the right hand side in Eqs.~\eqref{eq:better_off_more_ben_cvx_ar} and ~\eqref{eq:better_off_more_ben_cvx_tpr} represents constant terms since $\thetab_{sqo}$ is already known.

Both of the proposed proxies are convex with respect to the optimization variables. The convexity of the proxies (\ref{eq:better_off_more_ben_cvx_ar} and \ref{eq:better_off_more_ben_cvx_tpr}) means that for any convex function $L(\thetab)$ the optimization problem stays convex and can be solved in an efficient manner.

\xhdr{Logistic Regression: SP} We can specialize Problem~\eqref{eq:better_off_non_cvx}, using logistic regression classifier with L-2 norm regularizer, SP as a notion of discrimination, given by Eq.~\eqref{eq:sp_cvx}, and loss-averse constraint, given by Eq.~\eqref{eq:better_off_more_ben_cvx_tpr}, as follows: 

\begin{align}\label{eq:loss_averse_logreg_sp}
&\mbox{minimize}\quad - \frac{1}{|\Dcal|} \sum_{(\xb,y)\in\Dcal} \log p(y | \xb, \thetab) + \lambda ||\thetab||^2 \tag{P3} 
\\ \nonumber
&\mbox{subject to} \quad \frac{1}{|\Dcal|}
\bigg | 
\sum_{(\xb,z) \in \Dcal}  (z - \bar{z}) d_{\thetab}(\xb_i) 
\bigg| < c \\ \nonumber 
&\frac{1}{|\Dcal_{z=k}|} \sum_{\xb \in \Dcal_{z=k}} d_{\thetab}(\xb) \geq \frac{1}{|\Dcal_{z=k}|} \sum_{\xb \in \Dcal_{z=k}} d_{\thetab_{sqo}}(\xb) + \gamma,\\
&\quad \text{for all } k \in \{0,1\}, \gamma\in\RR^+. \nonumber
\end{align}
 \begin{figure*}[t]
 \centering
            \subfloat[Statistical Parity (SP)]
    {
    \includegraphics[angle=0, width=0.92\columnwidth]{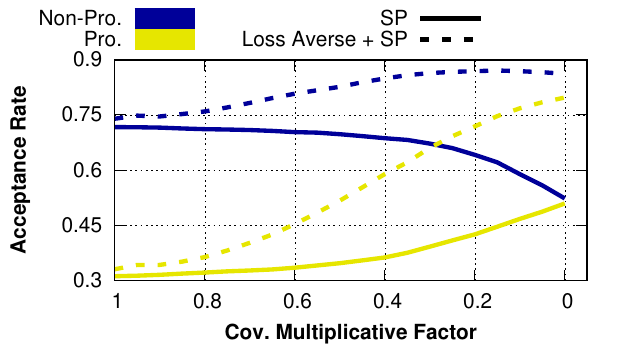}
    \label{fig:syn_ar_better_off_ar}
    }
            \subfloat[Nondiscrimination-accuracy tradeoff]
    {
    \includegraphics[angle=0, width=0.92\columnwidth]{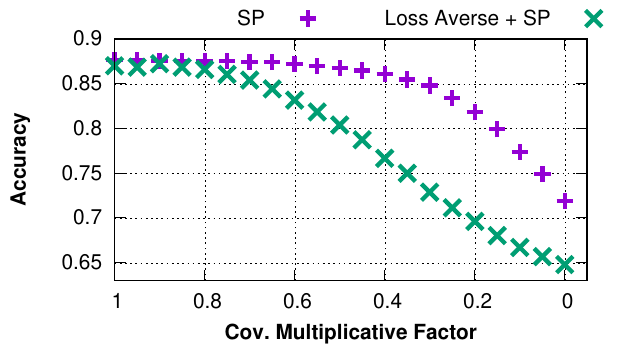}
    \label{fig:syn_ar_better_off_tradeoff}
    }
        \caption{
    [Synthetic dataset. Enforcing \stp{}] These figures show a comparison between the solutions of Problem~\eqref{eq:nondiscriminatory_clf}, using SP proxies, and Problem~\eqref{eq:loss_averse_logreg_sp}. Left panel shows the beneficial outcome rates, \ie, positive class acceptance rates, for a classifier only enforcing SP constraint (solid lines), and a classifier additionally enforcing the ``loss-averse'' constraint (dotted lines). Right panel shows the nondiscrimination-accuracy tradeoff for both the classifiers. Enforcing ``loss-averse'' constraint, defined in Eq.~\eqref{eq:better_off_more_ben_cvx_ar}, leads to significant additional loss in accuracy for the same level of discrimination.
    }
        \label{fig:synth_better_ar}
    \end{figure*}
\xhdr{Logistic Regression: EOP} Similarly, considering \eop{} as a notion of nondiscrimination we can approximate Problem~\eqref{eq:better_off_non_cvx}, by adding Eqs.~(\ref{eq:eop_cvx} and \ref{eq:better_off_more_ben_cvx_tpr}) as constraints to logistic loss,  as follows:
\begin{align}\label{eq:loss_averse_logreg_eop}
&\mbox{minimize}\quad - \frac{1}{|\Dcal|} \sum_{(\xb,y)\in\Dcal} \log p(y | \xb, \thetab) + \lambda||\thetab||^2 \tag{P4} 
\\ \nonumber
&\mbox{subject to} \quad \frac{1}{|\Dcal_{+}|}~
\bigg | 
\sum_{(\xb,z) \in \Dcal_{+}}  (z - \bar{z}) d_{\thetab}(\xb_i) 
\bigg| < c \\ \nonumber 
&\frac{1}{|\Dcal_{z=k}^{+}|} \sum_{\xb \in \Dcal_{z=k}^{+}} d_{\thetab}(\xb) \geq \frac{1}{|\Dcal_{z=k}^{+}|} \sum_{\xb \in \Dcal_{z=k}^{+}} d_{\thetab_{sqo}}(\xb) + \gamma,\\
&\quad \text{for all } k \in \{0,1\}, \gamma\in\RR^+. \nonumber
\end{align}

\section {Evaluation on Synthetic Dataset}\label{sec:eval}
In this section we evaluate the effectiveness ``Loss-averse'' constraint~\eqref{eq:better_off_more_ben_cvx_ar}, using a synthetic dataset on a binary classification task. We consider a well known notion of nondiscrimination, namely \stp{}. Due to space considerations, we show the results of loss-averse formulation, given by Eq.~\eqref{eq:better_off_more_ben_cvx_tpr}, combined with \eop{}, using synthetic data in Appendix~\ref{app:synth}. 
\subsection {Dataset and Experimental Set up}
We used synthetic dataset with binary ground truth class labels $y \in \{+1, -1\}$. Each data point comprises of $2$ features besides a binary sensitive feature, \ie, $z \in \{0, 1\}$, where $z = 0$ is the protected group. We do not use the sensitive attribute during training. 

\xhdr{Synthetic Dataset} For demonstrating the results of loss-averse updates with \stp, given by Eq.~\eqref{eq:sp_nonconvex}, as a notion of nondiscrimination, we used the dataset proposed by \citet{zafar_fairness}. This dataset comprises of 
$6000$ data points, the class labels were drawn uniformly at random.
Conditioned on the class membership, each data point was sampled from the following distributions:
\begin{align}
p(\xb | y = 1) &= N([2;2][5, 1;1, 5]),\nonumber \\
p(\xb | y = -1) &= N([-2;-2][10, 1;1, 3]).\nonumber 
\end{align}
Value of the sensitive attribute was sampled from the following Bernoulli probability distributions:

\begin{equation}
p(\zb = 1) = \frac{p(\xb^{'} | y = 1)}{p(\xb^{'} | y = 1) + p(\xb^{'} | y = -1)} ,\nonumber 
\end{equation}
where, $\xb^{'} = [\cos(\phi),- \sin(\phi); \sin(\phi), \cos(\phi)]\xb$, \ie, the rotated feature vector, $\xb$. On average there were $3280$ points in the protected group and $2720$ were in non-protected group.

\xhdr{Experimental Setup} The dataset is split into $70\%\text{-}30\%$, train-test folds. Additionally, hyperparameters are validated using a $30\%$ hold out set from the training data. All the results have been averaged over 5 shuffles of the data initialized by different random seed. 
In order to pick the penalization parameter, $\lambda$ in Problem~\eqref{eq:loss_averse_logreg_sp}, multiplied with the regularizer, we trained the unconstrained classifier for $\lambda \in [1e-5,1e-2]$. Then, we picked a value which yielded the highest accuracy on the validation set, for a particular shuffle of the data . We used this value of the parameter for {\it all} the experiments on that shuffle of the data. We use CVXPY~\cite{cvxpy} library to solve all the optimization problems.
\subsection {Loss-aversively Fair Updates}
In this section we experiment with Problems~(\ref{eq:nondiscriminatory_clf} and \ref{eq:loss_averse_logreg_sp}). First we consider \stp{}, where beneficial outcome rates are defined as positive class acceptance rate, as a notion of discrimination, \ie, solving Problems~\eqref{eq:nondiscriminatory_clf} using SP proxies. Then, we show results combining SP and loss-averse constraints and we update $\thetab_{sqo}$ with loss-averse nondiscriminatory classifiers.

\xhdr{Training Loss-aversively Fair Classifier} We initialize $\theta_{sqo}$ with the solution of unconstrained problem. Then, given a value of covariance threshold $c$, as used in Eqs.(\ref{eq:sp_cvx} and \ref{eq:eop_cvx}), and a range of $\gamma$, as used in Eqs.(\ref{eq:better_off_more_ben_cvx_ar} and \ref{eq:better_off_more_ben_cvx_tpr}), we solve Problem~\eqref{eq:loss_averse_logreg_sp}. We, then, pick the gamma values whose solutions yield a higher benefits compared to $\thetab_{sqo}$, for all the groups, on the validation set. In case there are multiple such values, we pick the one whose solution yields maximum accuracy. We then report the results on the test set. 

\xhdr{SP} Accuracy of an unconstrained classifier, on \textit{Synthetic} dataset, is $88\%$, and the acceptance rates for the protected and non-protected groups are $31\%$ and $72\%$, respectively. There is a clear disparity in acceptance rates of both the groups. In order to remove this disparity we solve Problem~\eqref{eq:nondiscriminatory_clf}, replacing the first constraint with SP proxy, given by Eq.~\eqref{eq:sp_cvx}. For a covariance threshold $c = 0$, this leads to a classifier with an acceptance rate of $51\%$ and $52\%$, for protected and non-protected groups respectively, and an accuracy of $72\%$. 

\noindent
The results for this formulation, Problem~\eqref{eq:nondiscriminatory_clf} specialized with SP, are shown in Figure~\eqref{fig:synth_better_ar}. The x-axis is covariance multiplicative factor $m: c = m \times c^* $, where $c^*$ is the covariance values of the unconstrained classifier and $c$ is covariance threshold as given in Eq.\eqref{eq:sp_cvx}. \textit{Solid lines} in Figure~\eqref{fig:syn_ar_better_off_ar} represent the statistics of the classifiers resulting from the solutions of this formulation. Figure~\eqref{fig:syn_ar_better_off_tradeoff} shows the accuracies of classifiers resulting from solving this formulation in \textit{purple} colored points.

\noindent
Note that: i) Figure~\eqref{fig:syn_ar_better_off_tradeoff} demonstrates that as the covariance is decreased the accuracy of the resulting, less discriminatory, classifiers also decreases. ii) Figure~\eqref{fig:syn_ar_better_off_ar} shows that as the covariance decreases, the discrimination also reduces. iii) However it should be noted that discrimination is decreased by reducing the acceptance rate of the non-protected group. 
\begin{figure*}[t]
 \centering
            \subfloat[Statistical Parity (SP)]
    {
    \includegraphics[angle=0, width=0.92\columnwidth]{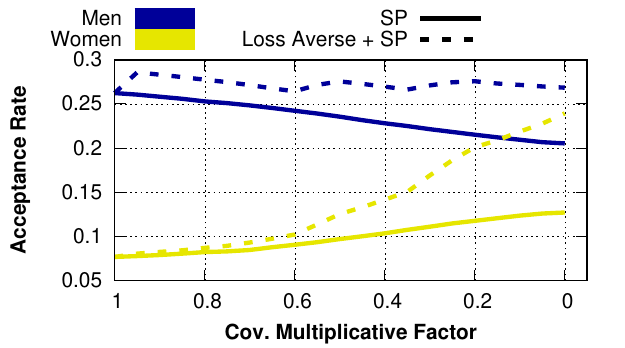}
    \label{fig:adult_ar_better_off_ar}
    }
            \subfloat[Nondiscrimination-accuracy tradeoff]
    {
    \includegraphics[angle=0, width=0.92\columnwidth]{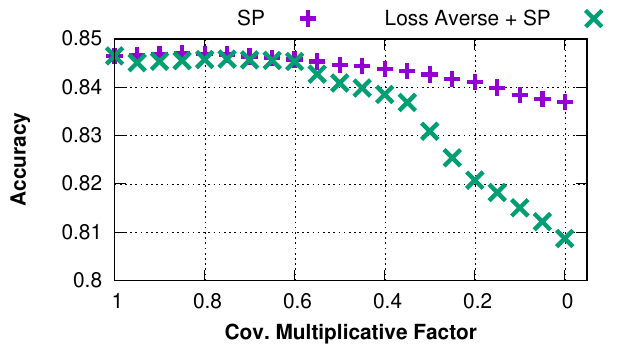}
    \label{fig:adult_ar_better_off_tradeoff}
    }
        \caption{
    [Adult dataset. Enforcing \stp{}] Left panel shows the beneficial outcome rates, \ie, positive class acceptance rates, for a classifier only enforcing SP constraint, \ie, solution of Problem~\eqref{eq:nondiscriminatory_clf} using SP proxies (solid lines), and a classifier additionally enforcing the ``loss-averse'' constraint, \ie, solution of Problem~\eqref{eq:loss_averse_logreg_sp} (dotted lines). Right panel shows the nondiscrimination-accuracy tradeoff for both the classifiers. Enforcing ``loss-averse'' constraint, defined in Eq.~\eqref{eq:better_off_more_ben_cvx_ar}, leads to a significant additional loss in accuracy for the same level of discrimination.
    }
        \label{fig:adult_better_ar}
    \end{figure*}

\xhdr{Loss-Aversiveness + SP} In order to train a classifier enforcing loss-averse  update of $\thetab_{sqo}$, Eq.~\eqref{eq:loss_averse_cons}, combined with statistical parity, Eq~\eqref{eq:sp_nonconvex}, on the \textit{Synthetic} dataset, we solve Problem~\eqref{eq:loss_averse_logreg_sp}. Loss-averse updates yield a classifier with an accuracy of $65\%$ and acceptance rates of $80\%$ and $86\%$ for protected and non-protected groups, respectively, for the covariance value $c = 0$. 

\noindent
The results are shown in Figure~\eqref{fig:syn_ar_better_off_ar} in \textit{dotted lines} and in \textit{green} colored points in Figure~\eqref{fig:syn_tpr_better_off_tradeoff}. i) The figures demonstrate that loss-aversively fair updates yield a less discriminatory classifier while increasing the benefits for both the groups, ii) however this comes at a higher cost of accuracy.

\xhdr{Summary} In this section we demonstrated the effectiveness of our proposed formulation on synthetic datasets. We illustrated the effectiveness of loss-aversively making the status quo classifiers nondiscriminatory, albeit at a higher cost of accuracy. 
\section {Evaluation on Real-World Dataset} \label{sec:eval_2}
In this section, we evaluate the effectiveness of our proposed schemes in updating the status quo classifier, $\thetab_{sqo}$, compliant with the ``loss-aversively fair updates'' consideration, on real-world dataset using \stp{} as a notion of nondiscrimination. We also consider another widely used notion of discrimination, \ie{}, \eop{}, and show loss-averse constraints combined with EOP on a real-world dataset in Appendix~\ref{app:real}, due to space limitations.

\subsection{Dataset and Experimental Setup}
In this section we explain the real-world dataset used to evaluate our proposed considerations.

\xhdr{Adult Dataset} We show result for loss-aversively fair update mechanism, introduced in section~\ref{sec:methodology}, using \textit{Adult dataset}~\cite{adult_dataset}. Specifically, we illustrate the effectiveness of Problem~\eqref{eq:loss_averse_logreg_sp} to train loss-aversively fair classifiers, using Adult dataset. For experiments in this section, we consider \stp{} as a notion of nondiscrimination.

The \textit{Adult Dataset} consists of $45,222$ subjects and 14 features like gender, race, educational level, \etc{}
The classification task is to predict whether a person earns more than 50K USD per annum (positive class) or not (negative class). We consider gender to be a sensitive feature for this dataset.

\xhdr{Experimental Setup} For the experiments conducted on the \textit{Adult dataset} we use the same data split as used for \textit{Synthetic dataset}. We also randomize the data, as well as validate the hyperparameters in a similar manner.

\subsection{Loss-Aversively Fair Updates} \label{sec:eval_better_off}
In this section we compare the results of Problem~\eqref{eq:nondiscriminatory_clf}, using SP proxies, and Loss-aversively fair updates given by Problem~\eqref{eq:loss_averse_logreg_sp} using \textit{Adult dataset}.  

\xhdr{SP} On the Adult dataset, logistic regression classifier leads to an accuracy of $84.6\%$. However, the classifier leads to the beneficial outcome rates of $8\%$ and $26\%$ for women and men respectively, showing a clear disparity in the beneficial outcome rates for the two groups.
Next, using the method of Zafar et al.~\cite{zafar_fairness}, we train a nondiscriminatory classifier while reducing the value of the covariance threshold $c$, (Eq.~\eqref{eq:sp_cvx}), towards 0. The results are shown in \textit{solid lines} in Figure~\eqref{fig:adult_ar_better_off_ar} and in \textit{purple} colored points in Figure~\eqref{fig:adult_ar_better_off_tradeoff}.
The least discriminatory classifier in this case achieves the beneficial outcome rates of $13\%$ and $20\%$ for women and men respectively, with an accuracy of $83.7\%$. We notice that the discrimination is reduced by lowering the beneficial outcome rates for men, which leads to a violation of ``loss-averse'' consideration.

\xhdr{Loss-Aversiveness + SP}
We next train classifier with the loss-averse constraints (Eq.~\eqref{eq:better_off_more_ben_cvx_ar}) combined with SP, \ie, solve Problem.~\eqref{eq:loss_averse_logreg_sp}. The least discriminatory classifier in this case achieves the beneficial outcome rates of $24\%$ and $27\%$ for women and men, respectively, while achieving an accuracy of $80.8\%$. However, the reduction in discrimination is achieved by only increasing the beneficial outcome rate for both groups. 
Results are shown in Figures~(\ref{fig:adult_ar_better_off_ar} and \ref{fig:adult_ar_better_off_tradeoff}), in \textit{dotted lines} and \textit{green} colored points, respectively.

\noindent
The figure shows the beneficial outcome rates for (i) a classifier with \stp{} constraint and (ii) a classifier with loss-averse and \stp{} constraints.  The figure shows that at successively decreasing values of the covariance threshold $c$, while classifier (i) achieves lower discrimination by increasing benefits for one group and decreasing them for the other, classifier (ii) does so by \emph{only increasing benefits for both the groups}. Figure~\eqref{fig:adult_ar_better_off_tradeoff} shows the nondiscrimination-accuracy tradeoff achieved by both the classifiers. The figure demonstrates that, as expected, classifier (ii) incurs a much higher cost in terms of accuracy for the same level of discrimination due to the additional loss-averse constraint.

\xhdr{Summary} Our proposed methodology, in Section~\ref{sec:methodology}, successfully enforces the loss averse constraint while updating the status quo classifier, $\thetab_{sqo}$, to a nondiscriminatory classifier. However, enforcing these constraints could be at a significant additional cost in terms of accuracy.

\section{Concluding Discussion} \label{sec:conclusion}

\noindent A number of recent works have explored various aspects of
fairness related to algorithmic decision making. In this paper, we
focus on an aspect of decision making that crucially affects people's
fairness perceptions, yet has been overlooked: it is the {\it fairness
  of updating decision making}, i.e., how the decision outcomes change
when updating a decision making system. 

Based on observations in behavioral economics and psychology, we note
that any ``disadvantageous'' changes in outcomes to
individual subjects or groups of subjects would be perceived as
unfair. Accordingly, we propose a complementary notion of update
fairness that we call {\it loss-averse updates}. Loss-averse updates try to constrain updates to only yield more advantageous (more beneficial) outcomes compared to status quo.

In this work, we formalize this notion in the context of
classification tasks. We proposed measures that would allow these
notions to be incorporated in the training of any convex
decision-boundary based classifiers (like logistic regression or
linear/non-linear SVM) as convex constraints. We also show how this
notion can be combined with prior notions and measures of
non-discrimination in classification. Our evaluation using synthetic and real-world datasets demonstrates the benefits of loss-averse updates in practice.

Our work here also opens up a number of new and interesting research
directions. The motivation behind our notions of fair updates
generalize to any algorithmic decision making scenario that affects
people's lives including search and recommender algorithms such as
Google's search, Facebook's NewsFeed, Amazon's product recommendations
or market-matching algorithms like Uber's rider-driver matching
algorithms. Exploring how our notion loss-averse updates can be applied to these more complex algorithmic decision making scenarios (beyond binary classification) remains an open challenge.

\bibliographystyle{ACM-Reference-Format}
\newpage
\bibliography{loss_aversively_fair}
\section*{Acknowledgements}
This research was supported in part by a European Research Council (ERC) Advanced Grant for the project “Foundations for Fair Social Computing”, funded under the European Union’s Horizon 2020 Framework Programme (grant agreement no. 789373)
\newpage

\begin{appendix}
\section{Evaluation on Synthetic Dataset: EOP} \label{app:synth}
\begin{figure*}[t]
 \centering
                \subfloat[Equality of Opportunity (EOP)]
    {
    \includegraphics[angle=0, width=0.92\columnwidth]{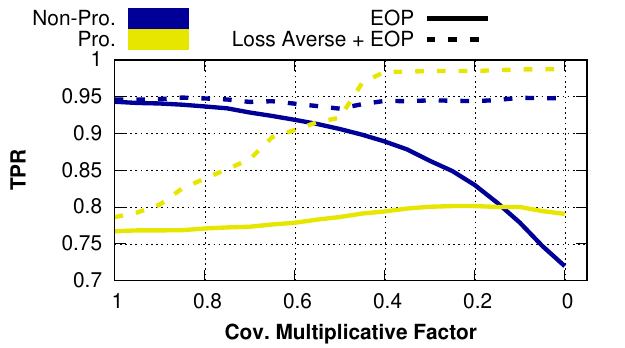}
    \label{fig:syn_tpr_better_off}
    }
            \subfloat[Nondiscrimination-accuracy tradeoff]
    {\includegraphics[angle= 0, width=0.92\columnwidth]
    {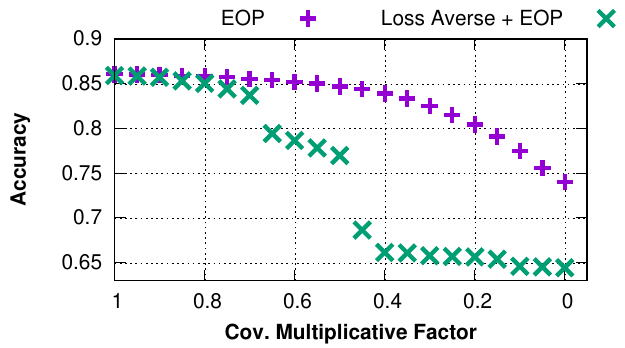}
    \label{fig:syn_tpr_better_off_tradeoff}
    }
            \caption{
    [Synthetic dataset. Enforcing \eop{}] Figure on the left shows the beneficial outcome rates, \ie, true positive rates, for a classifier only enforcing EOP constraint (solid lines) and a classifier additionally enforcing the ``loss-averse'' constraint, given in Eq.~\eqref{eq:better_off_more_ben_cvx_tpr}, is shown in dotted lines. Figure on the right shows nondiscrimination-accuracy tradeoff for both the classifiers.
    }
        \label{fig:synth_better_tpr}
    \end{figure*}
In this section we will present the ``loss-averse'' fairness results combined with \eop{}, using synthetic dataset. We show the results of the optimization Problem~\eqref{eq:loss_averse_logreg_eop}.
\subsection{Dataset and Experimental Setup}
In this section we explain the synthetic dataset used for demonstrating the loss-averse consideration and the experimental setup used to solve the optimization Problem~\eqref{eq:loss_averse_logreg_eop}.

\xhdr{Synthetic Dataset} Each data point comprises of 2 features apart from the sensitive attribute. Each data point also has a binary ground truth label. For \eop, as given by Eq.~\eqref{eq:eop_nonconvex}, we are considering true positive rates as a notion of benefit. To demonstrate the results of fair updates combined with EOP, we use a synthetic dataset proposed by \citet{zafar_dmt}, except that we flip the ground truth labels in order to have a disparity in the false negative rates instead of the false positive rates. We generated $16000$ data points with the probability distributions of the features given as follows:
\begin{align}
p(\xb | z = 0, y = 1) &= N([2;2][3, 1;1, 3])\nonumber \\
p(\xb | z = 1, y = 1) &= N([2;2][3, 1;1, 3])\nonumber \\ 
p(\xb | z = 0, y = -1) &= N([1;1][3, 3;1, 3])\nonumber \\
p(\xb | z = 1, y = -1) &= N([-2;-2][3, 1;1, 3])\nonumber
\end{align}
Both, class labels, $y$, and value of the sensitive attribute, $z$, were sampled uniformly at random.

\xhdr{Experimental Setup} We use the same data split and method of validating the hyperparameters as explained in section~\ref{sec:eval}. 

\subsection{Loss-Aversively Fair Updates}
In this section we show the results of Problem~\eqref{eq:nondiscriminatory_clf}, using EOP as a notion of nondiscrimination. We also show results for the loss-averse formulation combined with EOP, given by Problem~\eqref{eq:loss_averse_logreg_eop}.  

\xhdr{EOP} An unconstrained classifier trained on \textit{Synthetic} dataset yields an accuracy of $86\%$ and true positive rates (TPRs) of $94\%$ and $77\%$ for non-protected and protected groups, respectively. To equalize the TPRs we solve Problem~\eqref{eq:nondiscriminatory_clf} using proxies for EOP given in Eq.~\eqref{eq:eop_cvx}. 

\noindent
These results are show in Figure~\eqref{fig:syn_tpr_better_off} in \textit{solid lines} and Figure~\eqref{fig:syn_tpr_better_off_tradeoff} in \textit{purple} colored points. i) In order to reduce discrimination, this formulation yields a classifier which lowers the TPR of the non-protected class to $72\%$ and raises the TPR of the protected group to $79\%$, for covariance threshold $ c = 0$, while achieving an accuracy of $74\%$. ii) Figure~\eqref{fig:syn_tpr_better_off} shows the limitation of \eop{} proxy proposed by Zafar et al.~\cite{zafar_dmt}, as it achieves a lower discrimination for higher value of the covariance.

\xhdr{Loss-Aversiveness + EOP} To avoid lowering the benefits for any group while reducing discrimination, we solve the Problem~\eqref{eq:loss_averse_logreg_eop}. We encountered some issues in convergence for some values of covariance factor, specifically smaller ones. Out of $7$ random seeds that we tried we find the results for \textit{all} covariance factors for only $5$ seeds, we report the average of these results. One reason for the lack of convergence could be that a very high base TPR might make it difficult to find a nondiscriminatory classifier.For covariance threshold $c = 0$, this formulation leads to a classifier whose true positive rates are $95\%$ and $99\%$ for non-protected and protected groups, respectively, with an accuracy of $64\%$. 

We show these results in Figure~\eqref{fig:syn_tpr_better_off} in \textit{dotted lines} and Figure~\eqref{fig:syn_tpr_better_off_tradeoff} in \textit{green} colored crosses. i) These figures illustrate the effectiveness of the loss-averse formulation, as the resulting classifiers achieve nondiscrimination by increasing TPR for both groups, ii) however this results in a significant drop in the accuracy.

\section{Evaluation on Real-World Dataset: EOP} \label{app:real}
\begin{figure*}[t]
 \centering
            \subfloat[Equality of Opportunity (EOP)]
    {
    \includegraphics[angle=0, width=0.92\columnwidth]{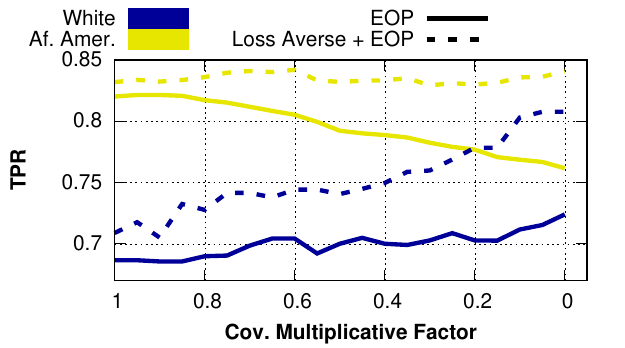}
    \label{fig:saf_better_off_tpr}
    }
            \subfloat[Nondiscrimination-accuracy tradeoff]{\includegraphics[angle= 0, width=0.92\columnwidth]
    {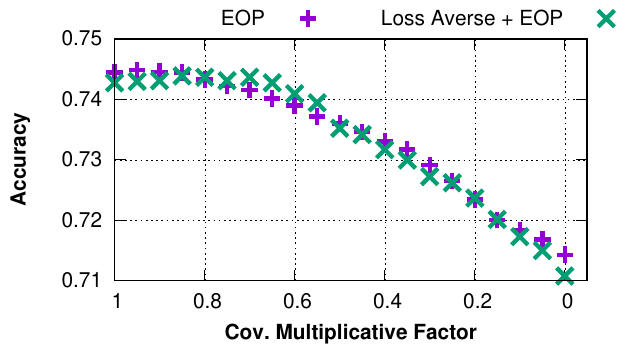}
    \label{fig:saf_better_off_tradeoff_tpr}
    }
        \caption{
    [SQF dataset. Enforcing \eop{}] 
    These figures show similar results as Figure~\eqref{fig:synth_better_tpr} using SQF dataset.
        }
        \label{fig:saf_better_tpr}
    \end{figure*}
In this section we will present the ``loss-averse'' fairness results combined with \eop{}, using a real-world dataset. 
\subsection{Dataset and Experimental Setup} In this section we explain the dataset and the experimental setup. We show result of
Problem~\eqref{eq:nondiscriminatory_clf}, with EOP as a notion of nondiscrimination, as well as Problem~\eqref{eq:loss_averse_logreg_eop}, which combines EOP and loss-averse constraints. 

\xhdr{SQF Dataset} For experiments in this section we consider \textit{ NYPD SQF dataset} \cite{sqf_dataset}. The NYPD \textit{SQF dataset} consists of pedestrians who were stopped in the year 2012 on the suspicion of having a weapon. The task is a binary prediction task which indicates whether (negative class) or not (positive class) a weapon was discovered. For our analysis, we consider the race to be the sensitive feature with values African-American and white. After balancing the classes and considering same features as Goel et al.~\cite{goel_frisk}, with the exception that we exclude the highly sparse features ‘precinct’ and ‘timestamp of the stop’, we obtain 5,832 subjects and 19 features.

\xhdr{Experimental Setup} We used similar experimental setup as explained in section~\ref{sec:eval}. 

\subsection{Loss-Aversively Fair Updates}
In this section we show the results of Problem~\eqref{eq:loss_averse_logreg_eop}, which enforces EOP and loss-averse constraints and compare them with the results of Problem~\eqref{eq:nondiscriminatory_clf}, which only enforces EOP using the proxy given by Eq.~\eqref{eq:eop_cvx}, on \textit{ NYPD SQF dataset}.

\xhdr{EOP}
With \eop{} constraint, where beneficial outcome rates are defined in terms of true positive rate, we experiment with NYPD SQF dataset.
Unconstrained logistic regression on SQF yields an accuracy of $74.4\%$, while the beneficial outcome rates are $69\%$ and $82\%$ for Whites and African-Americans, respectively. Least discriminatory classifier, trained with $c=0$, given in constraint Eq.~\eqref{eq:eop_cvx}, yields benefits of $72\%$ and $76\%$ for Whites and African-Americans, respectively, while achieving an accuracy of $71.4\%$. Similar to the previous cases, this classifier also achieves lower discriminations by raising the benefits for one group while increasing them for the other group.

\xhdr{Loss-Aversiveness + EOP}
Next, we combine the nondiscrimination constraint with the loss-averse constraint, given by Problem~\eqref{eq:loss_averse_logreg_eop}, in order to update $\thetab_{sqo}$. A least discriminatory loss-averse classifier trained on NYPD SQF dataset yields an accuracy of $71\%$ and benefits of $84\%$ and $81\%$ for African-Americans and White, respectively. 
Figure~\eqref{fig:saf_better_off_tpr} shows the beneficial outcome rates for (i) a classifier with only nondiscrimination constraints and (ii) a loss-averse classifier with nondiscrimination constraints. 
Again, we notice that classifier (ii) removes discrimination by \emph{only increasing the beneficial outcome rates} whereas classifier (i) does so by increasing benefits for one group and decreasing them for the other. Finally, the comparison of nondiscrimination-accuracy tradeoff in Figure~\eqref{fig:saf_better_off_tradeoff_tpr} shows no significant difference between both the classifiers.

\end{appendix}
\end{document}